\documentclass[runningheads]{llncs}

\usepackage{eccv}

\usepackage{eccvabbrv}

\usepackage{graphicx}
\usepackage{booktabs}

\usepackage[accsupp]{axessibility}  %

\usepackage[dvipsnames]{xcolor}
\newcommand{\red}[1]{{\color{black}#1}}

\usepackage[utf8]{inputenc}
\usepackage[T1]{fontenc}

\usepackage{url}
\usepackage{booktabs}
\usepackage{multirow}
\usepackage{multicol}
\usepackage{amsfonts}
\usepackage{nicefrac}
\usepackage{microtype}
\usepackage{xcolor}
\usepackage{placeins}	
\usepackage{graphicx}
\usepackage{subcaption}
\usepackage{algorithm2e}
\usepackage{enumitem}

\usepackage{pifont}
\newcommand{\cmark}{\ding{51}}
\newcommand{\xmark}{\ding{55}}

\usepackage[percent]{overpic}
\usepackage{tcolorbox}

\usepackage{fontawesome5}  %
\usepackage{capt-of}  %

\newcommand{\faSearchPluss}{\reflectbox{\faSearchPlus}}
\newcommand{\faSearchMinuss}{\reflectbox{\faSearchMinus}}

\tcbset{standard jigsaw, on line,
	boxsep=1.5pt, left=0pt,right=0pt,top=0pt,bottom=0pt,
	opacityback=0.8,opacityframe=.8,
	colframe=white,
	colback=white
}

\usepackage[pagebackref,breaklinks,colorlinks
,citecolor=eccvblue
]{hyperref}

\usepackage{orcidlink}

\begin{document}

\title{Data Upcycling Knowledge Distillation \\ for Image Super-Resolution} 

\titlerunning{DUKD}

\author{\hspace{-0.4cm}Yun Zhang$^{1,2}$ ~ Wei Li$^{2}$ ~ Simiao Li$^{2}$ ~ Hanting Chen$^{2}$\\ 
    \hspace{-0.4cm}Zhijun Tu$^{2}$ ~ Wenjia Wang$^{1}$ ~ Bingyi Jing$^{3}$ ~ Shaohui Lin$^{4}$ ~ Jie Hu$^{2}$\thanks{Corresponding author: hujie23@huawei.com}\\
	$^{1}$Hong Kong University of Science and Technology (GZ)\\
	$^{2}$Huawei Noah's Ark Lab\\
	$^{3}$Southern University of Science and Technology\\
    $^{4}$East China Normal University\\
	\tt{\hspace{0cm}{yzhangjy@connect.ust.hk}}\quad\quad {\tt\{wei.lee, lisimiao\}@huawei.com}
}
\institute{}
\authorrunning{Y. Zhang et al.}

\maketitle

\begin{abstract}  %
Knowledge distillation (KD) compresses deep neural networks by transferring task-related knowledge from cumbersome pre-trained teacher models to compact student models. %
\red{
However, current KD methods for super-resolution (SR) networks overlook the nature of SR task that the outputs of the teacher model are noisy approximations to the ground-truth distribution of high-quality images (GT), which shades the teacher model's knowledge to result in limited KD effects.
}
To utilize the teacher model beyond the GT upper-bound, we present the Data Upcycling Knowledge Distillation (DUKD), to transfer the teacher model’s knowledge to the student model through the upcycled in-domain data derived from training data. %
Besides, we impose label consistency regularization to KD for SR by the paired invertible augmentations to improve the student model's performance and robustness.
Comprehensive experiments 
demonstrate that the DUKD method significantly outperforms previous arts on several~SR~tasks. 

\end{abstract}
    
\section{Introduction}\label{sec: introduction}

Image super-resolution~(SR) is a class of fundamental but challenging tasks within the realm of computer vision~(CV), aiming to reconstruct high-resolution~(HR) images from their low-resolution~(LR) counterparts~\cite{lim2017enhanced, zhang2018image, liang2021swinir}. Over the past decade, convolutional neural networks~(CNNs)~\cite{dong2015image, kim2016accurate,lim2017enhanced,zhang2018image} and Transformers~\cite{liang2021swinir, yang2020learning, wang2022uformer,zamir2022restormer,zhang2024distilling,qiao2024lipt,wang2023ift} have demonstrated remarkable success for SR. Despite the impressive performance of deep learning-based SR models, their practical deployment is typically hindered by the requirement of high computational resources and memory footprint~\cite{zhang2021aligned}. Consequently, there has been a growing interest in developing SR model compression methods to facilitate real-world applications particularly on resource-limited devices~\cite{Huang_2023_CVPR}. 

Knowledge Distillation~(KD) reduces the computational costs and memory requirements for practical model deployments while also considerably improving the performance of the student models. It's realized by transferring ``dark knowledge'' from the well-performing but cumbersome teacher model to the lightweight student model~\cite{gao2019image, hui2019lightweight, zhang2021data}. Compared to other model compression methods such as quantization~\cite{gupta2015deep, hubara2016binarized,ignatov2021real,wu2016quantized}, pruning~\cite{anwar2017structured, wang2021exploring,liu2019metapruning,qiao2022dcs} and the neural architectures search~(NAS)~\cite{wu2019fbnet,howard2019searching,guo2020single}, KD has received intensive attention due to its excellent performance and broad applicability.

\begin{figure*}[t]  %
	\centering
	\includegraphics[width=\linewidth]{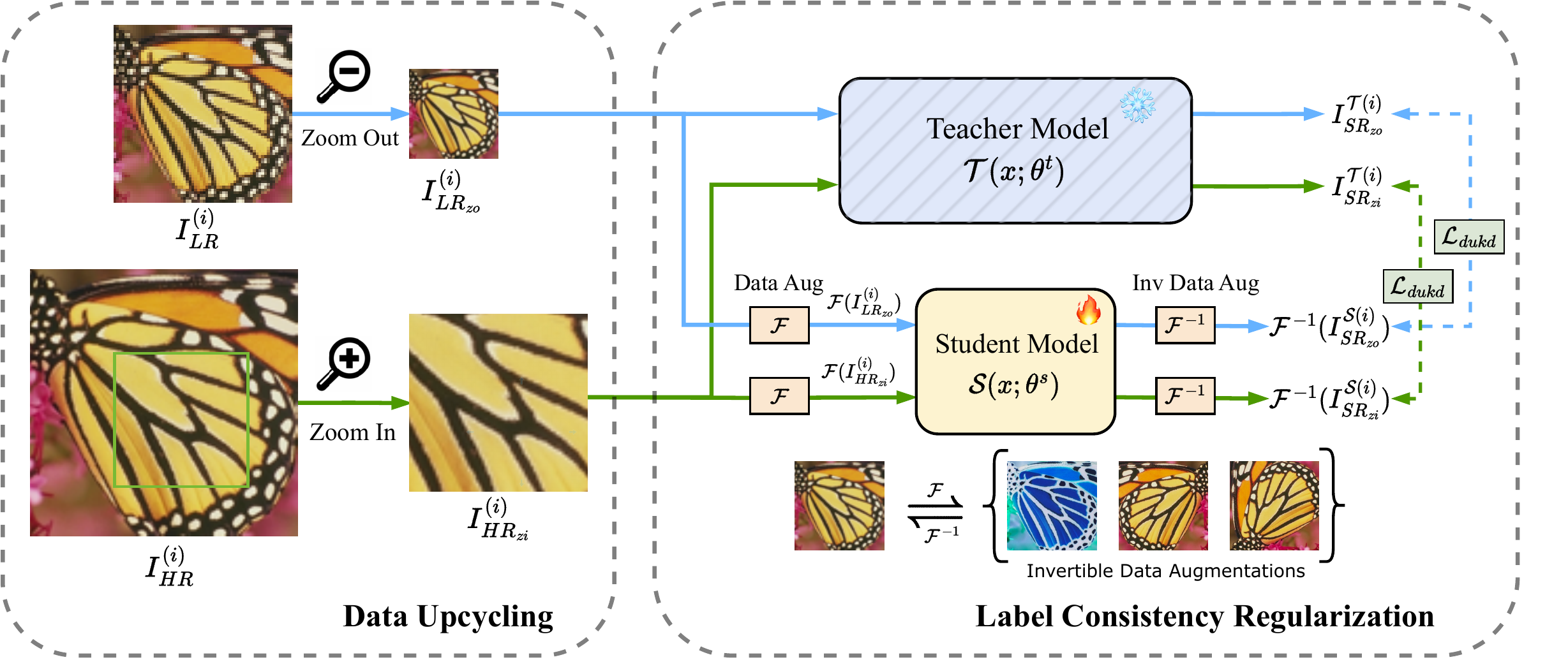}
	\caption[Distillation Framework]{Framework of the DUKD method. It facilitates the student with the prior knowledge provided by the teacher through upcycled in-domain data. The label consistency regularization enhances the generalizability of the student. Besides $\mathcal{L}_{dukd}$, the total loss also includes the conventional $\mathcal{L}_{rec}$ and $\mathcal{L}_{kd}$, which are omitted for simplicity.
 }
\label{fig:framework}\vspace{-1em}
\end{figure*}

The efficacy of KD has been well-established in natural language processing (NLP)~\cite{gou2021knowledge,sanh2019distilbert} and high-level CV tasks such as classification, detection and segmentation~\cite{park2019relational,tung2019similarity,chen2017learning,huang2017like}. However, its application in SR tasks is comparatively less researched~\cite{he2020fakd, wang2021towards, zhang2021data,lee2020learning,li2024knowledge}. The utilization of vanilla response-based KD methods~\cite{hinton2015distilling} or those that are effective on high-level CV tasks~\cite{romero2014fitnets, yim2017gift, zagoruyko2016paying} yields marginal improvements or may even suffer from detrimental effects when used to distil SR networks, as noted by He~\textit{et al.}~\cite{he2020fakd} and our experiments in~\cref{sec: experiments}.
Prior KD methods tailored explicitly for SR are primarily feature-based, 
forcing the student model to mimic the teacher model's network intermediate features directly~\cite{he2020fakd} or through a pre-trained perceptual feature extracting network like VGG~\cite{Yao2022MTKDSRMK, wang2021towards}. %
However, these feature-based approaches have marginal performance gains and limited applicability. Practically, the teacher models' architecture is often invisible due to commercial, privacy, and safety concerns, rendering feature-based methods invalid in some real-world applications.

\par 
Investigating the mechanism of KD in the context of SR, we found that many of the previous explanations for why KD works no longer hold due to the unique characteristics of SR. Since the teacher model's output, as a noisy approximation to the GT image, contains barely extra information over GT, so the ``dark knowledge'' of teacher model are hardly transferred to student model through KD. 
Different from exist approaches, we investigate the knowledge inside training data for the SR task and present the the \textit{Data Upcycling Knowledge Distillation (DUKD)}, a simple yet highly effective KD framework for efficient SR. 
\cref{fig:framework} illustrates the process of DUKD. 
It shifts the paradigm from developing various knowledge types~\cite{gou2021knowledge} to more task-adapted training data mining and construction with the aid of pre-trained teacher. 
Specifically, the DUKD consists of two major modules: \textit{in-domain data upcycling} and \textit{label consistency regularization}. 
The in-domain data upcycling module utilizes the training pairs to build auxiliary training examples which are used by teacher model to teach the student model. 
It frees the teacher model from being an inaccurate repeater of the GT labels.
Moreover, we realize the label consistency regularization into the KD for SR by defining several \textit{invertible data augmentation} operations. 
The student model is forced to yield the same output as the teacher model, given the enhanced inputs. The regularization makes the student model exposed to a diverse range of inputs, substantially improving performance~\cite{oliver2018realistic, jeong2019consistency, englesson2021consistency}.
The DUKD is independent of network architectures, and shows great universality among a diverse array of SR model families and SR tasks.
In summary, our main contributions are three-fold:
\begin{itemize}
	\item[$\bullet$] We analyze the mechanisms of KD for SR networks, and propose DUKD based on the unique properties of SR models.  %
    \item[$\bullet$] We leverage the label consistency regularization into KD for SR by specifying several invertible data augmentations. It improves the model's robustness to perturbed input images.
	\item[$\bullet$] The proposed DUKD, 
 applies broadly to multiple teacher-student configurations, promising a cutting-edge KD approach for SR. %
 Our code\footnote{Experiment code is available on GitHub: \href{https://github.com/yun224/DUKD}{https://github.com/yun224/DUKDi}} is is publicly available.
\end{itemize}

\section{Related Works}\label{sec: related}
\subsection{Image Super-Resolution}
Deep neural networks~(DNNs) based image super-resolution have shown impressive success. Dong \textit{et al.}~\cite{dong2014learning} firstly introduced CNN with only three convolution layers for image SR. Then, residual learning was introduced in VDSR~\cite{kim2016accurate}, reaching 20 convolution layers. Lim \textit{et al.} proposed EDSR~\cite{lim2017enhanced} with a simplified residual blocks~\cite{ResNet}. Zhang \textit{et al.} proposed an even deeper network, RCAN~\cite{zhang2018image}. Most of them have achieved state-of-the-art results with deeper and wider networks. 
Recently, Transformer has gained a surge of interest in image restoration. Chen \textit{et al.} proposed the first pre-trained image processing transformer IPT~\cite{chen2021pre}. 
SwinIR~\cite{liang2021swinir} applies the Swin-Transformer architecture to the image restoration for deep feature extraction.
Restormer~\cite{zamir2022restormer} proposed a multi-scale hierarchical design incorporating efficient Transformer blocks by modifying self-attention and MLP. 
While CNNs and Transformers have demonstrated impressive performance in SISR, they suffer from high memory and computational costs.

\subsection{Knowledge Distillation}
\textbf{KD for high-level CV}. 
Knowledge distillation is widely recognized as an effective model compression method that can significantly reduce the computation overload and improve student's capability~\cite{hinton2015distilling, yim2017gift, gou2021knowledge}. The response-based KD methods are simple yet effective where the student models directly imitate the predictions or logits of the teacher model~\cite{hinton2015distilling,zhao2022decoupled,chen2017learning}.
The proposed DUKD method falls into this category since only the final outputs of models are aligned.
Besides the output of the networks, the intermediate features can also be used to improve the student model, by matching feature maps directly,  after dimension alignment by algebra operations~\cite{zagoruyko2016paying} or extra modules~\cite{kim2018paraphrasing,passban2021alp,guo2021distilling}. 
The relations between layers or samples are also used for KD, such as correlation~\cite{yim2017gift,you2017learning}, mutual information~\cite{passalis2020heterogeneous}, and pairwise or triple-wise geometric relations~\cite{park2019relational}.

\noindent\textbf{KD for low-level CV}. 
Lately, there has been an increasing number of efforts made on the KD for super-resolution networks. 
He \textit{et al.} proposed the FAKD to align the dimensions of models' feature maps by spatial affinity matrix to train the student model~\cite{he2020fakd}. Lee \textit{et al.} employed an encoder to extract the compact features from HR images to initialize the generator network and thereby perform feature distillation~\cite{lee2020learning}. Wang \textit{et al.} proposed a channel-sharing self-distillation method with perceptual contrastive losses~\cite{wang2021towards}. To train SR network under the privacy and data transmission limitations, Zhang \textit{et al.} employed a generator to support data-free KD~\cite{zhang2021data}.
The common limitation of these methods is that they are only applicable to CNN-based models and have certain requirements on the teacher-student structure. %

\section{Methodology}\label{sec: dukd}

\subsection{Notations and Preliminaries}
Let $\mathcal{T}(x;\theta^{t})$ and $\mathcal{S}(x;\theta^{s})$ be a teacher and a student SR model with parameters $\theta^{t}$ and $\theta^{s}$ for the super-resolution of input $x$, respectively. Given an input pair of LR and HR image $I_{LR}^{(i)}\in\mathbb{R}^{H\times W\times 3}$, $I_{HR}^{(i)}\in\mathbb{R}^{s_cH\times s_cW\times 3}$, the output SR images of the two networks are denoted by $I_{SR}^{\mathcal{T}(i)} = \mathcal{T}(I_{LR}^{(i)};\theta^{t})$ and $I_{SR}^{\mathcal{S}(i)} = \mathcal{S}(I_{LR}^{(i)};\theta^{s})$, where $H\times W$ is the input size and $s_c\in\mathbb{Z}^{+}$ is the scaling factor.
The L1-norm reconstruction loss is computed as:
\vspace{-0.25em}\begin{equation}\label{eq:rec-loss}
	\mathcal{L}_{rec} = \|I_{SR}^{\mathcal{S}(i)} - I_{HR}^{(i)}\|_1.
\end{equation}
And the vanilla response-based KD loss is given by
\vspace{-0.25em}\begin{equation}\label{eq:kd-loss}
	\mathcal{L}_{kd} = \|I_{SR}^{\mathcal{S}(i)} - I_{SR}^{\mathcal{T}(i)}\|_1,
\end{equation}
which is directly from the output of teacher and student models.

\begin{figure}[htb]
	\centering
	\includegraphics[width=0.9\columnwidth]{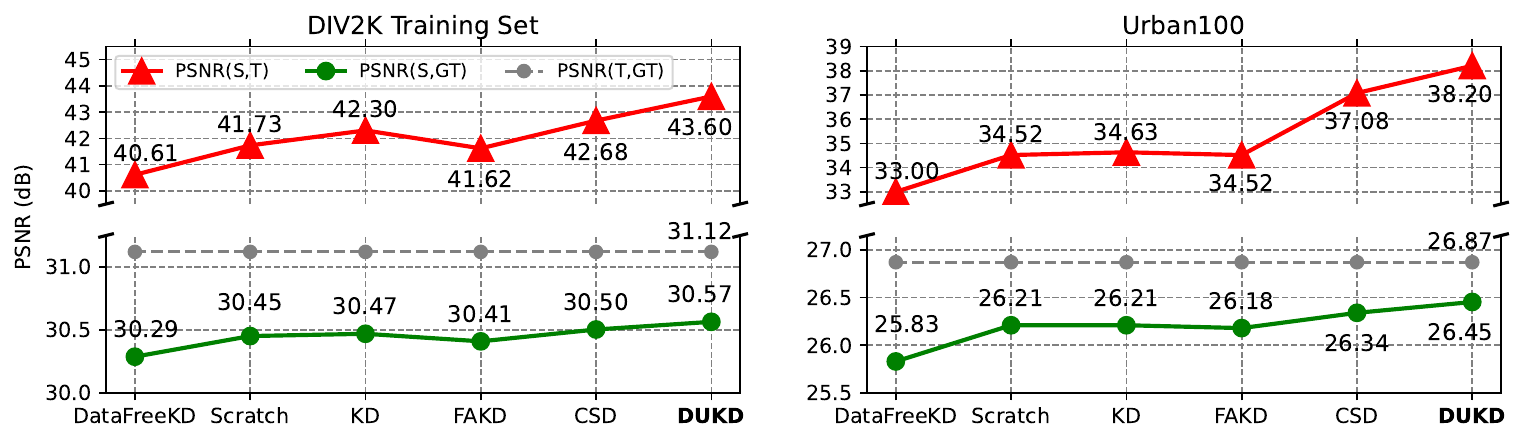}%
	\caption{Similarity between the student and teacher \texttimes4 EDSR models over different training approaches (indicated by the x-axis labels). \texttt{PSNR(S,T)} denotes the average PSNR between student and teacher models'~outputs, with larger values reflecting higher similarity. \texttt{PSNR(S,GT)} denotes the average PSNR between the output of the student model and ground-truth HR image, with larger values showing better fitting (left: training set) or generalization performance (right: testing set).
	}\vspace{-1.5em}
	\label{fig:fedility-generalization}
\end{figure}

\subsection{Motivation} \label{sec: motivation}

Since the emergence of the knowledge distillation technique~\cite{hinton2015distilling}, several analyses and discussions have been carried out on the mechanism of why teacher supervision contributes to improving the performance of student models~\cite{tang2020understanding,stanton2021does,wang2021revisiting,zhang2022quantifying,harutyunyan2023supervision}. It has been widely agreed that for response-based KD, dark knowledge from the teacher model are the inter-class and inter-examples relation information contained in the output logits, which is absent in the ground-truth labels.

\par
\begin{figure}[t]
    \centering
    \includegraphics[width=\linewidth]{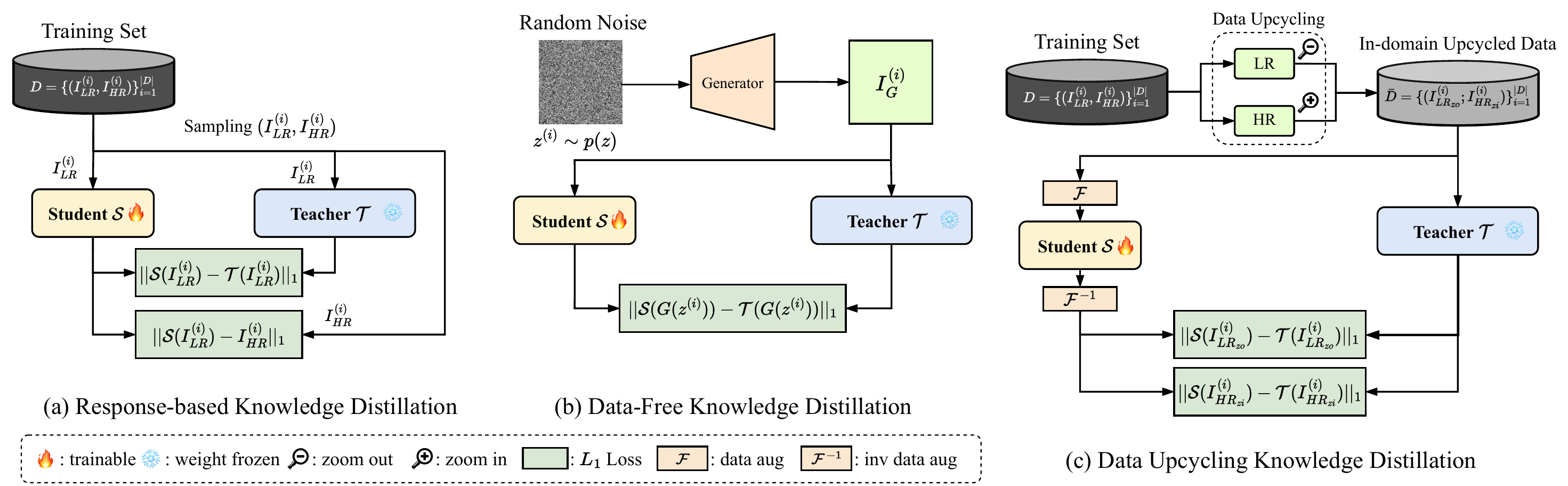}
    \caption{Comparison between the logits-KD, Data Free KD, and DUKD.}\vspace{-1.5em}
    \label{fig:logits-datafree-dukd}
\end{figure}

 However, there is barely such benefit in SR tasks that reconstructs image pixels. Since the outputs of SR network $I_{SR}^{T(i)}$ are noisy and inaccurate approximations to the ground-truth distribution of high-resolution image $I_{HR}^{(i)}$, as shown in~\cref{fig:logits-datafree-dukd}~(a). Directly aligning the model outputs hardly transfers knowledge to and may even mislead the student model. 
The teacher model's distribution information is shaded by $I_{HR}^{(i)}$, resulting in limited KD effects. 
To verify this hypothesis, we train a \texttimes4 scale EDSR network by different training methods (Data-free KD, No-distillation, Logits-KD, FAKD, CSD, and the proposed DUKD). To make the models comparable, the data-free KD uses the LR from training set and discards HR, by assuming there is an oracle generator $G$. Then we compute the PSNR metrics between the outputs of teacher and student models, on the training and testing sets respectively, which reflect the similarities between networks. 
The results shown in~\cref{fig:fedility-generalization} indicate that the existing KD approaches make the student model predicts more like teacher only to a small extent, since the the improvements on \texttt{PSNR(S,T)} over no-distillation (train from scratch) are limited. Therefore, the PSNR referring to GT, \texttt{PSNR(S,GT)}, are also low on both training and testing sets.

\par This issue cannot be addressed by data augmentation since it just reuses the available image pairs, and the ``recycled'' data are inadequate to differ the function of teacher model from GT supervision.
The data-free knowledge distillation methods~\cite{zhang2021data} stay out of this problem to some extent due to their abandon of references from the training data. The supervision signals solely come from the teacher model, as illustrated in~\cref{fig:logits-datafree-dukd} (b). 
Although teacher model's knowledge are transferred to student model, it's unreasonable to discard the training data especially when they are available. Besides, the teacher model may yield more noisy output on the generated images.
Above analyses and findings motivate us to build more task-adapted KD framework by mining information from the training data. Specifically, we construct auxiliary training inputs for functioning KD, which are closely related with the training set to prevent distribution shift.

\subsection{Data Upcycling} \label{sec: unlabeled-data}
The overall framework of DUKD is demonstrated in Fig.\ref{fig:framework}. It showcases how the in-domain upcycled data examples are constructed to support KD.
The data upcycling module consists of two image zooming operations %
. The zoom-in~\faSearchPluss~operation is facilitated by randomly cropping a patch from $I_{HR}^{(i)}$, which has the same size as the LR image $I_{LR}^{(i)}$ for the convenience of batch processing. Conversely, the zoom-out~\faSearchMinuss~operation is carried out by down-sampling the LR image in the way exactly the same as how $I_{LR}^{(i)}$ is generated from $I_{HR}^{(i)}$. 
For a pair of training examples ($I_{LR}^{(i)}$, $I_{HR}^{(i)}$), the output of the zoom-out operation is unique, but the zoom-in operation on $I_{HR}^{(i)}$ could result in various outcomes according to the strategy of patch selection. Beyond random cropping, regions can also be selected based on their reconstruction difficulty or texture complexity. However, it's observed in our experiments that adapted selection would incur a higher computational cost with marginal performance gains.

After the zooming operations, the teacher model outputs corresponding SR images for the zoom-in and zoom-out images to supervise the student model. 
Since there is only the teacher model's supervision for this portion of data, 
its distribution information are unshaded from GT and able to impact the student model. 
DUKD provides a more refined and data-centric approach to KD, reflecting its benefits in effective data utilization and superior performance in SR tasks.
The overall loss is constructed by adding the extra DUKD loss term that is computed on the upcycled data to the reconstruction loss (Eq. \ref{eq:rec-loss}) and conventional KD loss (Eq. \ref{eq:kd-loss}). For the input $(I_{LR}^{(i)}, I_{HR}^{(i)})$ image pair, denote the upcycled data as $I_{LR_{zo}}^{(i)}, I_{LR_{zi}}^{(i)}$,
\begin{equation}\label{key}%
	\mathcal{L}_{dukd} = \|I_{SR_{zo}}^{\mathcal{S}(i)} - I_{SR_{zo}}^{\mathcal{T}(i)}\|_1 + \|I_{SR_{zi}}^{\mathcal{S}(i)} - I_{SR_{zi}}^{\mathcal{T}(i)}\|_1,
\end{equation}
where $I_{SR_{zo}}^{\mathcal{S}(i)} = \mathcal{S}(I_{LR_{zo}}^{(i)};\theta^{s})$, $I_{SR_{zo}}^{\mathcal{T}(i)} = \mathcal{T}(I_{LR_{zo}}^{(i)};\theta^{t})$ and the other terms are computed similarly. If zoom-out is performed, we compute the reconstruction loss between $I_{SR_{zo}}^{\mathcal{S}(i)}$ and $I_{LR}^{(i)}$ as well. To sum up, 
\begin{equation}
\mathcal{L} = \mathcal{L}_{rec} + \lambda_{kd} \mathcal{L}_{kd} + \lambda_{dukd} \mathcal{L}_{dukd},
\end{equation}
where $\lambda_{kd}$ and $\lambda_{dukd}$ are the loss weights.

\red{
\subsection{Label Consistency Regularization}

\begin{figure*}[t]  %
	\centering
	\includegraphics[width=0.925\linewidth]{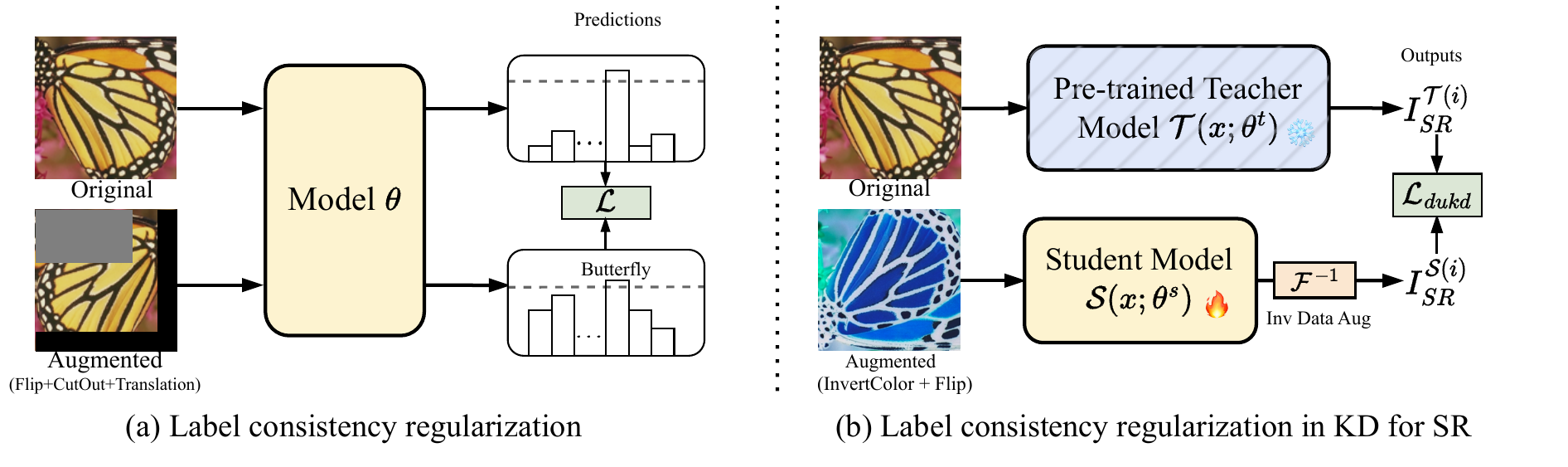}
	\caption[Label Consistency Regularization]{Comparison of the label consistency regularization in high-level CV and KD for SR. The augmentations should be invertible to make the models' output comparable.}
\label{fig:label-consistency-regularization}\vspace{-1em}
\end{figure*}

Consistency regularization is commonly used in semi-supervised and self-supervised learning. As illustrated in~\cref{fig:label-consistency-regularization} (a), it encourages the prediction of the network to be consistent over perturbed training examples, leading to robustness against corrupted data in test time~\cite{oliver2018realistic, englesson2021consistency,jeong2019consistency}. 
The model is expected to correctly identify the crucial semantic information related to specific tasks from the input, despite the possible noise and perturbations. 
This regularization is based on various image augmentation techniques, such as rotation, shearing, cutout, and translation.

Given the input data, knowledge distillation forces the  student model to output the same predictions as the teacher model. Such characteristic should be retained for augmented input since the task-related semantic information remains unchanged and the input perturbations should not significantly distinguish between the outputs of the teacher and student models.
To realize label consistency regularization, we impose data augmentations on the input of student model while leaving the original input for the teacher model. It would ideally result in the student model learning invariant processes from diverse transformations. Meanwhile, the student is supervised by a more powerful teacher model, whose supervision are from non-perturbed inputs that inherently possess superior quality compared to those derived from augmented ones. 
Thereby the upcycled data and the teacher model are better leveraged.
Taking the upcycled zoom-in image as an example,  
the consistency regularization can be represented as:
\vspace{-0.25em}\begin{equation*}
    \mathcal{L} = ||\mathcal{S}(\mathcal{F}(I_{HR_{zi}}); \theta^s) - \mathcal{T}(I_{HR_{zi}}; \theta^t)||_1 
    ,
\end{equation*}
where $\mathcal{F}(\cdot)$ denotes the perturbation function.

However, as SR is a pixel-level image-to-image CV task that is weakly relevant to semantic information of image subject, any tweak on the input can alter the model's output. For KD, the student model's output would consequently be incomparable with the teacher model's. Therefore, we need to perform inverse augmentation, namely $\mathcal{F}^{-1}(\cdot)$, on the output of the student model. The label consistency regularization becomes:
\vspace{-0.25em}\begin{equation*}
    \mathcal{L} =  || \mathcal{F}^{-1}(\mathcal{S}(\mathcal{F}(I_{HR_{zi}}); \theta^s)) - \mathcal{T}(I_{HR_{zi}}; \theta^t)||_1 
    .
\end{equation*}
}
The selected augmentations should be invertible and relevant to the SR task for maintaining the crucial pixel-level details after augmentation. It requires that for any input image $I$, $\mathcal{F}^{-1}(\mathcal{F}(I))=I$.
Hence, a number of popular image augmentations, such as blurring, cutout, brightness adjustment, and cropping, are not applicable as they do not meet this prerequisite.
Instead, we employ two geometric transformations, horizontal/vertical flip and 90{\textdegree}/180{\textdegree}/270{\textdegree} rotations, along with a novel \textit{color inversion} transformation 
that subtracts each pixel intensity value of the input image from 255 (or 1 if normalized): $\mathcal{F}(I) = 255 - I$.
The color inversion is invertible and maintains the relative magnitude among pixel values. It also prompts the student models to be more sensitive to essential structural features such as lines and edges.
Right bottom of~\cref{fig:framework} illustrates the three types of invertible data augmentations employed to realize label consistency.

\begin{figure*}[t]
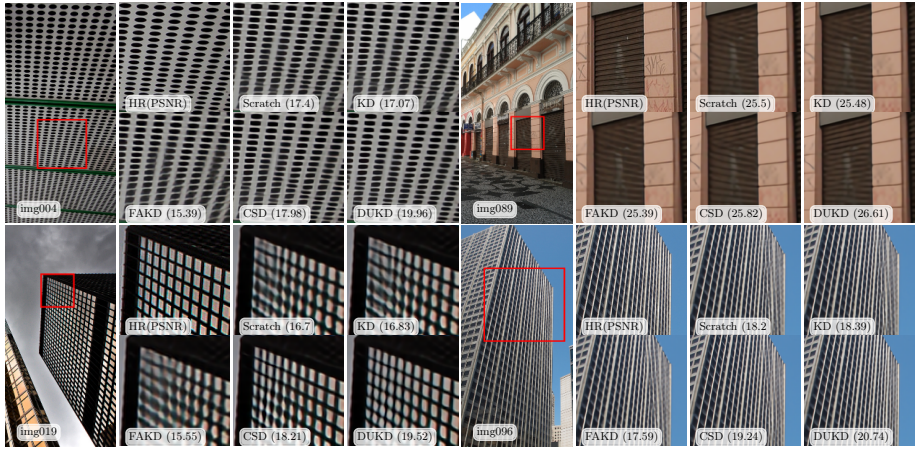

		\resizebox{\textwidth}{!}{\hspace{0.1em}
			\begin{minipage}{0.25\linewidth}
				\begin{overpic}[width=\textwidth]{figure/visualization/img004\_gt\_v\_cr}
					\put(5,5){\tcbox{img004}}
				\end{overpic}
			\end{minipage}
			\begin{minipage}{0.25\linewidth}
				\begin{overpic}[width=\textwidth]{figure/visualization/img004\_gtsub}
					\put(5,5){\tcbox{HR(PSNR)}}
				\end{overpic}
				\begin{overpic}[width=\textwidth]{figure/visualization/img004\_FAKD}
					\put(5,5){\tcbox{FAKD (15.39)}}
				\end{overpic}
			\end{minipage}
			\begin{minipage}{0.25\linewidth}
				\begin{overpic}[width=\textwidth]{figure/visualization/img004\_Scratch}
					\put(5,5){\tcbox{Scratch (17.4)}}
				\end{overpic}
				\begin{overpic}[width=\textwidth]{figure/visualization/img004\_CSD}
					\put(5,5){\tcbox{CSD (17.98)}}
				\end{overpic}
			\end{minipage}
			\begin{minipage}{0.25\linewidth}
				\begin{overpic}[width=\textwidth]{figure/visualization/img004\_KD}
					\put(5,5){\tcbox{KD (17.07)}}
				\end{overpic}
				\begin{overpic}[width=\textwidth]{figure/visualization/img004\_Ours}
					\put(5,5){\tcbox{DUKD (19.96)}}
				\end{overpic}
			\end{minipage}
			
			\begin{minipage}{0.25\linewidth}
				\begin{overpic}[width=\textwidth]{figure/visualization/img089\_gt\_v\_cr}
					\put(5,5){\tcbox{img089}}
				\end{overpic}
			\end{minipage}
			\begin{minipage}{0.25\linewidth}
				\begin{overpic}[width=\textwidth]{figure/visualization/img089\_gtsub}
					\put(5,5){\tcbox{HR(PSNR)}}
				\end{overpic}
				\begin{overpic}[width=\textwidth]{figure/visualization/img089\_FAKD}
					\put(5,5){\tcbox{FAKD (25.39)}}
				\end{overpic}
			\end{minipage}
			\begin{minipage}{0.25\linewidth}
				\begin{overpic}[width=\textwidth]{figure/visualization/img089\_Scratch}
					\put(5,5){\tcbox{Scratch (25.5)}}
				\end{overpic}
				\begin{overpic}[width=\textwidth]{figure/visualization/img089\_CSD}
					\put(5,5){\tcbox{CSD (25.82)}}
				\end{overpic}
			\end{minipage}
			\begin{minipage}{0.25\linewidth}
				\begin{overpic}[width=\textwidth]{figure/visualization/img089\_KD}
					\put(5,5){\tcbox{KD (25.48)}}
				\end{overpic}	
				\begin{overpic}[width=\textwidth]{figure/visualization/img089\_Ours}
					\put(5,5){\tcbox{DUKD (26.61)}}
				\end{overpic}
			\end{minipage}
		} %
		\resizebox{\textwidth}{!}{\hspace{0.1em}
			\begin{minipage}{0.25\linewidth}
				\begin{overpic}[width=\textwidth]{figure/visualization/img019\_gt\_v\_cr}
					\put(5,5){\tcbox{img019}}
				\end{overpic}
			\end{minipage}
			\begin{minipage}{0.25\linewidth}
				\begin{overpic}[width=\textwidth]{figure/visualization/img019\_gtsub}
					\put(5,5){\tcbox{HR(PSNR)}}
				\end{overpic}
				\begin{overpic}[width=\textwidth]{figure/visualization/img019\_FAKD}
					\put(5,5){\tcbox{FAKD (15.55)}}
				\end{overpic}
			\end{minipage}
			\begin{minipage}{0.25\linewidth}
				\begin{overpic}[width=\textwidth]{figure/visualization/img019\_Scratch}
					\put(5,5){\tcbox{Scratch (16.7}}
				\end{overpic}
				\begin{overpic}[width=\textwidth]{figure/visualization/img019\_CSD}
					\put(5,5){\tcbox{CSD (18.21)}}
				\end{overpic}
			\end{minipage}
			\begin{minipage}{0.25\linewidth}
				\begin{overpic}[width=\textwidth]{figure/visualization/img019\_KD}
					\put(5,5){\tcbox{KD (16.83)}}
				\end{overpic}
				\begin{overpic}[width=\textwidth]{figure/visualization/img019\_Ours}
					\put(5,5){\tcbox{DUKD (19.52)}}
				\end{overpic}
			\end{minipage}
			
			\begin{minipage}{0.25\linewidth}
				\begin{overpic}[width=\textwidth]{figure/visualization/img096\_gt\_v\_cr}
					\put(5,5){\tcbox{img096}}
				\end{overpic}
			\end{minipage}
			\begin{minipage}{0.25\linewidth}
				\begin{overpic}[width=\textwidth]{figure/visualization/img096\_gtsub}
					\put(5,5){\tcbox{HR(PSNR)}}
				\end{overpic}
				\begin{overpic}[width=\textwidth]{figure/visualization/img096\_FAKD}
					\put(5,5){\tcbox{FAKD (17.59)}}
				\end{overpic}
			\end{minipage}
			\begin{minipage}{0.25\linewidth}
				\begin{overpic}[width=\textwidth]{figure/visualization/img096\_Scratch}
					\put(5,5){\tcbox{Scratch (18.2}}
				\end{overpic}
				\begin{overpic}[width=\textwidth]{figure/visualization/img096\_CSD}
					\put(5,5){\tcbox{CSD (19.24)}}
				\end{overpic}
			\end{minipage}
			\begin{minipage}{0.25\linewidth}
				\begin{overpic}[width=\textwidth]{figure/visualization/img096\_KD}
					\put(5,5){\tcbox{KD (18.39)}}
				\end{overpic}	
				\begin{overpic}[width=\textwidth]{figure/visualization/img096\_Ours}
					\put(5,5){\tcbox{DUKD (20.74)}}
				\end{overpic}
			\end{minipage}
			
		}
	
	\caption{The \texttimes4 SR examples of EDSR models on img004, img019, img089 and img096 from Urban100. PSNRs (dB) of the cropped regions are annotated below each image.}\vspace{-1em}
	
	\label{fig:vis}
\end{figure*}

\subsection{Difference with Data Augmentation}
The DUKD framework diverges significantly from conventional data augmentation or expansion techniques in the following ways:

\par Firstly, the data upcycling procedure is intimately intertwined with KD, wherein the teacher model generates the HR label corresponding to the auxiliary training samples, namely the upcycled in-domain LR image. This distinct characteristic contrasts with data augmentations typically imposed on the original input data independent of teacher models. The augmented data is the ``recycling'' of the original data, and the ground-truth label still hinders functioning of KD.
Data upcycling also different from the dataset expansion that introduces extra out-domain data which may incur distribution shift issue. In DUKD, data upcycling is not merely an operation to increase the volume or diversity of the data but instead serves a more significant purpose - it acts as a data medium for the teacher model to transmit its knowledge to the student model without the interference of GT. 
This unique procedure allows the student model to learn more effectively from the teacher model's responses to the upcycled data, thus enhancing its performance.

\par Secondly, DUKD introduces the label consistency regularization to KD, a novel approach to improve the student model's robustness against perturbations. 
It allows the student model to learn the teacher's invariant features across transformations and encourages it to better generalize to diverse data. 
The label consistency regularization is realized by selected invertible data augmentations applied only on the student model's input and output. 
We use certain data augmentations as a tool to yield consistency.

\begin{table}[t]
	\caption{SR model specifications and statistics (\texttimes 4 scale). The FLOPs and frames per second (FPS) are computed with a 3\texttimes256\texttimes256 input image on single V100 GPU of 64GB VRAM. The block denotes the number of residual blocks for EDSR and RCAN  (in each residual group) or Swin transformer blocks for SwinIR models.}\vspace{-0.5em}
	\label{tab:model-config}
	\centering
	\renewcommand{\arraystretch}{0.8}
	\begin{tabular}{@{}llrrrrrr@{}}
		\toprule
		\multirow{2}{*}{Model} & \multirow{2}{*}{Role} & \multicolumn{3}{c}{Network}         & \multirow{2}{*}{FLOPs (G)} & \multirow{2}{*}{\#Params} & \multirow{2}{*}{FPS} \\ \cmidrule(lr){3-5}
		&                       & Channel & Block & Group &                            &                           &                      \\ \midrule
		\multirow{2}{*}{EDSR}  & Teacher               & 256       & 32         & -          & 3293.35                    & 43.09 M                   & 3.233                \\
		& Student               & 64        & 32         & -          & 207.28                     & 2.70 M                    & 33.958               \\ \midrule
		\multirow{2}{*}{RCAN}  & Teacher               & 64        & 20         & 10         & 1044.03                    & 15.59 M                   & 6.162                \\
		& Student               & 64        & 6          & 10         & 366.98                     & 5.17 M                    & 12.337               \\ \midrule
		\multirow{2}{*}{SwinIR} & Teacher               & 180        &   6       &    -      & 861.27                    & 11.90 M                   &    0.459             \\
		& Student               & 60        &      4     &   -       & 121.48                     & 1.24 M                    &    0.874            \\ \bottomrule
	\end{tabular}
\end{table}

\section{Experiments}\label{sec: experiments}
\subsection{Experiment Setups}\label{sec: exp-setting}

\textbf{Backbones and Baselines.} We use EDSR~\cite{lim2017enhanced}, RCAN~\cite{zhang2018image}, and SwinIR~\cite{liang2021swinir} as backbone models to verify the effectiveness of DUKD and compare it with existing KD methods. The specifications of teacher and student networks and some statistics including FLOPs, number of parameters, and model inference speed (FPS) are presented in~\cref{tab:model-config}. We compare DUKD with the following baseline training and KD methods: training from scratch, response-based KD~\cite{hinton2015distilling}, FitNet~\cite{romero2014fitnets}, AT~\cite{zagoruyko2016paying}, RKD~\cite{park2019relational}, FAKD~\cite{he2020fakd}, and CSD~\cite{wang2021towards}. The FitNet, AT, and RKD are originally proposed for high-level CV tasks, but they are compatible with SR task and applicable to CNN models. Since the CSD is a self-distillation method of channel-splitting manner, it's not applicable to the distillation of RCAN (network depth compression) and SwinIR (transformers architecture) models. Similarly, several KD methods are incompatible with SwinIR models due to their feature-based distillation characteristic.
To evaluate the performance of SR models, we compute the peak signal-to-noise ratio (PSNR) and the structural similarity index (SSIM) on the Y channel of the YCbCr color space conventionally.

\noindent\textbf{Training Details.} The SR models are trained with the 800 training images from DIV2K~\cite{timofte2017ntire} and evaluated on four benchmark datasets: Set5~\cite{bevilacqua2012low}, Set14~\cite{zeyde2012single}, BSD100~\cite{martin2001database}, and Urban100~\cite{huang2015single}. The LR images used for training were obtained by down-sampling the HR images with the bicubic degradation. The \texttimes 4 scale SR models are initialized with the corresponding \texttimes 2 ones. During training, the input LR image is randomly cropped into $48\times48$ patches and augmented by random horizontal and vertical flips and rotations. For FAKD and CSD methods, we follow the hyperparameters setting specified in original paper and train the models by ourselves if checkpoint is not provided, as particularly noted in the results table. The zoom in~\faSearchPluss~operation of DUKD is conducted by randomly cropping for simplicity. The zoom out~\faSearchMinuss~is skipped for training SwinIR since the~$I_{LR_{zo}}$~would be too small to be valid input for the model. All the models are trained with ADAM optimizer~\cite{kingma2014adam} with $\beta_1 = 0.9$, $\beta_2 = 0.99$ and $\epsilon = 10^{-8}$, with a batch size of 16 and a total of $2.5\times 10^{5}$ updates. The initial learning rate is set to $10^{-4}$ and is decayed by a factor of 10 at every $10^{5}$ iterations. We implemented the proposed kd method with the BasicSR~\cite{basicsr} and PyTorch 1.10 framework and trained models with 4 NVIDIA V100 GPUs.

\begin{table}[htb]
\caption{Quantitative comparison (average PSNR/SSIM) between DUKD and other distillation methods for \textbf{EDSR} of three SR scales. The best and second-best performances are highlighted in bold and underlined, respectively. The asterisk indicates that the results in a row are from our reproduction experiments.}\vspace{-0.5em}
\label{tab:exp-edsr}
\centering
\renewcommand{\arraystretch}{0.9}
\begin{tabular}{@{\hspace{2pt}}l@{\hspace{1.5\tabcolsep}}lcccc@{\hspace{2pt}}}
\toprule
\multirow{2}{*}[-0.24em]{Scale} &
  \multirow{2}{*}[-0.24em]{Method} &
  Set5 &
  Set14 &
  BSD100 &
  Urban100 \\ \cmidrule(l){3-6} 
 &
   &
  PSNR/SSIM &
  PSNR/SSIM &
  PSNR/SSIM &
  PSNR/SSIM \\ \midrule
\multirow{8}{*}{\texttimes2} &
  Scratch &
  38.00/0.9605 &
  33.57/0.9171 &
  32.17/0.8996 &
  31.96/0.9268 \\
 &
  KD &
  38.04/0.9606 &
  33.58/0.9172 &
  32.19/0.8998 &
  31.98/0.9269 \\
 &
  FitNet &
  37.59/0.9589 &
  33.09/0.9136 &
  31.79/0.8953 &
  30.46/0.9111 \\
 &
  AT &
  37.96/0.9603 &
  33.48/0.9167 &
  32.12/0.8990 &
  31.71/0.9241 \\
 &
  RKD &
  38.03/0.9606 &
  33.57/0.9173 &
  32.18/0.8998 &
  31.96/0.9270 \\
 &
  FAKD$^\ast$ &
  37.99/0.9606 &
  33.60/0.9173 &
  32.19/0.8998 &
  32.04/0.9275 \\
 &
  CSD$^\ast$ &
  \underline{38.06}/\underline{0.9607} &
  \underline{33.65}/\underline{0.9179} &
  \underline{32.22}/\underline{0.9004} &
  \underline{32.26}/\underline{0.9300} \\
 &
  \textbf{DUKD} &
  \textbf{38.15}/\textbf{0.9610} &
  \textbf{33.80}/\textbf{0.9195} &
  \textbf{32.27}/\textbf{0.9007} &
  \textbf{32.53}/\textbf{0.9320} \\ \midrule
\multirow{8}{*}{\texttimes3} &
  Scratch &
  34.39/0.9270 &
  30.32/0.8417 &
  29.08/0.8046 &
  27.99/0.8489 \\
 &
  KD &
  34.43/0.9273 &
  30.34/0.8422 &
  29.10/0.8050 &
  28.00/0.8491 \\
 &
  FitNet &
  33.35/0.9178 &
  29.71/0.8323 &
  28.62/0.7949 &
  26.61/0.8167 \\
 &
  AT &
  34.29/0.9262 &
  30.26/0.8406 &
  29.03/0.8035 &
  27.76/0.8443 \\
 &
  RKD &
  34.43/0.9274 &
  30.33/0.8423 &
  29.09/0.8051 &
  27.96/0.8493 \\
 &
  FAKD$^\ast$ &
  34.39/0.9272 &
  30.34/0.8426 &
  29.10/0.8052 &
  28.07/0.8511 \\
 &
  CSD$^\ast$ &
  \underline{34.45}/\underline{0.9275} &
  \underline{30.32}/\underline{0.8430} &
  \underline{29.11}/\underline{0.8061} &
  \underline{28.21}/\underline{0.8549} \\
 &
  \textbf{DUKD} &
  \textbf{34.59}/\textbf{0.9287} &
  \textbf{30.47}/\textbf{0.8448} &
  \textbf{29.20}/\textbf{0.8073} &
  \textbf{28.44}/\textbf{0.8578} \\ \midrule
\multirow{8}{*}{\texttimes4} &
  Scratch &
  32.29/0.8965 &
  28.68/0.7840 &
  27.64/0.7380 &
  26.21/0.7893 \\
 &
  KD &
  32.30/0.8965 &
  28.70/0.7842 &
  27.64/0.7382 &
  26.21/0.7897 \\
 &
  FitNet &
  31.65/0.8873 &
  28.33/0.7768 &
  27.38/0.7309 &
  25.40/0.7637 \\
 &
  AT &
  32.22/0.8952 &
  28.63/0.7825 &
  27.59/0.7365 &
  25.97/0.7825 \\
 &
  RKD &
  32.30/0.8965 &
  28.69/0.7842 &
  27.64/0.7383 &
  26.20/0.7899 \\
 &
  FAKD$^\ast$ &
  32.27/0.8960 &
  28.65/0.7836 &
  27.62/0.7379 &
  26.18/0.7895 \\
 &
  CSD &
  \underline{32.34}/\underline{0.8974} &
  \underline{28.72}/\underline{0.7856} &
  \underline{27.68}/\underline{0.7396} &
  \underline{26.34}/\underline{0.7948} \\
 &
  \textbf{DUKD} &
  \textbf{32.47}/\textbf{0.8981} &
  \textbf{28.80}/\textbf{0.7866} &
  \textbf{27.71}/\textbf{0.7403} &
  \textbf{26.45}/\textbf{0.7963} \\ \bottomrule
\end{tabular}%
\vspace{-0.5em}
\end{table}

\begin{table}[htb]
\caption{Quantitative comparison (average PSNR/SSIM) between DUKD and other distillation methods for \textbf{RCAN} of three SR scales. The best and second-best performances are highlighted in bold and underlined, respectively.}\vspace{-0.5em}
\label{tab:exp-rcan}
\centering
\renewcommand{\arraystretch}{0.9}
\begin{tabular}{@{\hspace{2pt}}l@{\hspace{1.5\tabcolsep}}lcccc@{\hspace{2pt}}}
\toprule
\multirow{2}{*}[-0.24em]{Scale} &
  \multirow{2}{*}[-0.24em]{Method} &
  Set5 &
  Set14 &
  BSD100 &
  Urban100 \\ \cmidrule(l){3-6} 
 &
   &
  PSNR/SSIM &
  PSNR/SSIM &
  PSNR/SSIM &
  PSNR/SSIM \\ \midrule
\multirow{7}{*}{\texttimes2} &
  Scratch &
  38.13/0.9610 &
  33.78/0.9194 &
  32.26/0.9007 &
  32.63/0.9327 \\
 &
  KD &
  38.18/0.9611 &
  33.83/0.9197 &
  32.29/0.9010 &
  32.67/0.9329 \\
 &
  FitNet &
  37.97/0.9602 &
  33.57/0.9174 &
  32.19/0.8999 &
  32.06/0.9279 \\
 &
  AT &
  38.13/0.9610 &
  33.70/0.9187 &
  32.25/0.9005 &
  32.48/0.9313 \\
 &
  RKD &
  \underline{38.18}/\underline{0.9612} &
  33.78/0.9191 &
  32.29/0.9011 &
  \underline{32.70}/\underline{0.9330} \\
 &
  FAKD$^\ast$ &
  38.17/0.9612 &
  \underline{33.83}/\underline{0.9199} &
  \underline{32.29}/\underline{0.9011} &
  32.65/0.9330 \\
 &
  \textbf{DUKD} &
  \textbf{38.23}/\textbf{0.9614} &
  \textbf{33.90}/\textbf{0.9201} &
  \textbf{32.33}/\textbf{0.9016} &
  \textbf{32.87}/\textbf{0.9349} \\ \midrule
\multirow{7}{*}{\texttimes3} &
  Scratch &
  34.61/0.9288 &
  30.45/0.8444 &
  29.18/0.8074 &
  28.59/0.8610 \\
 &
  KD &
  34.61/0.9291 &
  30.47/0.8447 &
  \underline{29.21}/\underline{0.8080} &
  \underline{28.62}/\underline{0.8612} \\
 &
  FitNet &
  34.21/0.9248 &
  30.20/0.8399 &
  29.05/0.8044 &
  27.89/0.8472 \\
 &
  AT &
  34.55/0.9287 &
  30.43/0.8438 &
  29.17/0.8070 &
  28.43/0.8577 \\
 &
  RKD &
  \underline{34.67}/\underline{0.9292} &
  30.48/0.8451 &
  29.21/0.8080 &
  28.60/0.8610 \\
 &
  FAKD$^\ast$ &
  34.63/0.9290 &
  \underline{30.51}/\underline{0.8453} &
  29.21/0.8079 &
  28.62/0.8612 \\
 &
  \textbf{DUKD} &
  \textbf{34.74}/\textbf{0.9296} &
  \textbf{30.54}/\textbf{0.8458} &
  \textbf{29.25}/\textbf{0.8088} &
  \textbf{28.79}/\textbf{0.8646} \\ \midrule
\multirow{7}{*}{\texttimes4} &
  Scratch &
  32.31/0.8966 &
  28.69/0.7842 &
  27.64/0.7384 &
  26.37/0.7949 \\
 &
  KD &
  32.45/0.8980 &
  28.76/0.7860 &
  27.67/0.7400 &
  26.49/0.7980 \\
 &
  FitNet &
  31.99/0.8899 &
  28.50/0.7789 &
  27.55/0.7353 &
  25.90/0.7791 \\
 &
  AT &
  32.31/0.8967 &
  28.69/0.7839 &
  27.64/0.7385 &
  26.29/0.7927 \\
 &
  RKD &
  32.39/0.8974 &
  28.74/0.7856 &
  27.67/0.7399 &
  26.47/0.7981 \\
 &
  FAKD$^\ast$ &
  \underline{32.46}/\underline{0.8980} &
  \underline{28.77}/\underline{0.7860} &
  \underline{27.68}/\underline{0.7400} &
  \underline{26.50}/\underline{0.7980} \\
 &
  \textbf{DUKD} &
  \textbf{32.56}/\textbf{0.8990} &
  \textbf{28.83}/\textbf{0.7870} &
  \textbf{27.72}/\textbf{0.7410} &
  \textbf{26.62}/\textbf{0.8020} \\ \bottomrule
\end{tabular}%
\end{table}

\begin{table}[tb]
\caption{Quantitative comparison (average PSNR/SSIM) between DUKD and other applicable distillation methods for \textbf{SwinIR} of three SR scales. Best performance is highlighted in bold.}\vspace{-0.5em}
\label{tab:exp-swinir}
\centering
\renewcommand{\arraystretch}{0.9}
\begin{tabular}{@{\hspace{2pt}}l@{\hspace{1.5\tabcolsep}}lcccc@{\hspace{2pt}}}
\toprule
\multirow{2}{*}[-0.24em]{Scale} &
  \multirow{2}{*}[-0.24em]{Method} &
  Set5 &
  Set14 &
  BSD100 &
  Urban100 \\ \cmidrule(l){3-6} 
 &
   &
  PSNR/SSIM &
  PSNR/SSIM &
  PSNR/SSIM &
  PSNR/SSIM \\ \midrule
\multirow{3}{*}{\texttimes2} &
  Scratch &
  38.01/0.9607 &
  33.57/0.9178 &
  32.19/0.9000 &
  32.05/0.9279 \\
 &
  KD &
  38.04/0.9608 &
  33.61/0.9184 &
  32.22/0.9003 &
  32.09/0.9282 \\
 &
  \textbf{DUKD} &
  \textbf{38.13}/\textbf{0.9610} &
  \textbf{33.78}/\textbf{0.9194} &
  \textbf{32.26}/\textbf{0.9007} &
  \textbf{32.63}/\textbf{0.9327} \\ \midrule
\multirow{3}{*}{\texttimes3} &
  Scratch &
  34.41/0.9273 &
  30.43/0.8437 &
  29.12/0.8062 &
  28.20/0.8537 \\
 &
  KD &
  34.44/0.9275 &
  30.45/0.8443 &
  29.14/0.8066 &
  28.23/0.8545 \\
 &
  \textbf{DUKD} &
  \textbf{34.55}/\textbf{0.9285} &
  \textbf{30.53}/\textbf{0.8456} &
  \textbf{29.20}/\textbf{0.8080} &
  \textbf{28.53}/\textbf{0.8604} \\ \midrule
\multirow{3}{*}{\texttimes4} &
  Scratch &
  32.31/0.8955 &
  28.67/0.7833 &
  27.61/0.7379 &
  26.15/0.7884 \\
 &
  KD &
  32.27/0.8954 &
  28.67/0.7833 &
  27.62/0.7380 &
  26.15/0.7887 \\
 &
  \textbf{DUKD} &
  \textbf{32.41}/\textbf{0.8973} &
  \textbf{28.79}/\textbf{0.7860} &
  \textbf{27.69}/\textbf{0.7405} &
  \textbf{26.43}/\textbf{0.7972} \\ \bottomrule
\end{tabular}%
\end{table}

\subsection{Results and Comparison}\label{sec: exp-results}

\noindent\textbf{Comparison with Baseline Methods. }
The quantitative results (PSNR / SSIM) for training EDSR, RCAN, and SwinIR networks are presented in~\cref{tab:exp-edsr}, \ref{tab:exp-rcan}, and \ref{tab:exp-swinir} respectively, for \texttimes 2, \texttimes 3, and \texttimes 4 scales. The following conclusions can be drawn from these results: (\textbf{1}) Existing KD methods have limited benefits and some even result in student models worse than those trained without KD. For instance, EDSR models trained with FAKD sometimes underperform the ones trained from scratch. (\textbf{2}) The KD methods originally designed for high-level CV tasks (FitNet, AT, RKD), while applicable, hardly improve the SR models over training from scratch. (\textbf{3}) The DUKD framework presented in this work consistently outperforms the existing KD baseline methods in all experimental settings by a large margin. For example, compared with the response-based KD method, the average PSNR improvements for the three types of networks on the Urban100 test set over three SR scales are 0.43 dB, 0.31 dB, 0.31 dB, respectively. Most existed KD methods are inapplicable to the transformer architecture network, but DUKD, as a response-based KD method, is able to compress the SwinIR model while exhibiting great performance.
\par\noindent\textbf{DUKD facilitates the student model to mimic the teacher model.} In Fig.~\ref{fig:fedility-generalization}, we demonstrate the effectiveness of different KD methods by comparing the similarity of students' output towards teacher's on the training and Urban100 testing sets to evaluate if the student learns well to mimic the teacher model. It shows that DUKD makes the student not only effectively fit the teacher model on the training set but also imitate it on the test sets so that the student model's outputs get closer to GT as well.

\begin{table}[htb]  %
\begin{minipage}[t]{0.4\linewidth}
	\centering
	\caption{The results of heterogeneous distillation using DUKD on the \texttimes 4 scale RCAN model. 
 }\vspace{-0.5em}
	\label{tab:heter-kd}
		\begin{tabular}{@{}c@{\hspace{3\tabcolsep}}c@{\hspace{3\tabcolsep}}c@{}}
			\toprule
			\multirow{2}{*}[-0.24em]{Teacher}   & BSD100 & Urban100 \\ \cmidrule(lr){2-3} 
			& PSNR/SSIM  & PSNR/SSIM   \\ \midrule
			(Scratch) & 27.64/0.7384 & 26.37/0.7949 \\ \midrule
			EDSR    & 27.71/0.7406 & 26.59/0.8014 \\
			SwinIR  & 27.72/0.7408 & 26.59/0.8007 \\ \bottomrule
		\end{tabular}%
\end{minipage}
\quad
\begin{minipage}[t]{0.525\linewidth}

\caption{NIQE scores on several real-world SR testing datasets. The lower, the better. Visual comparisons are provided in the appendix.}\vspace{-0.5em}
\label{tab:realworld-sr}
\centering
\resizebox{\columnwidth}{!}{%
\begin{tabular}{@{\hspace{2pt}}llccc@{\hspace{2pt}}}
\toprule
Method & \#Params & RealSR  & DRealSR    & OST300    \\ \midrule
Scratch                 & 11.9M                    & 4.771 	    & 4.847 	 & 2.932 
  \\ \midrule
Scratch                 & \multirow{3}{*}{1.24M}   & 5.810 	    & 5.757 	 & 3.788  \\
KD                      &                          & 5.425 	    & 5.408 	 & 3.652 
      \\
DUKD                    &                          & 5.398 	    & 5.378 	 & 3.494 
       \\ \bottomrule
\end{tabular}%
}
\end{minipage}
\end{table}

\noindent\textbf{Experiment Results on Heterogeneous Settings. }
We extend the experiments to heterogeneous settings where the teacher and student models have different network architectures, as presented in Tab. \ref{tab:heter-kd}. Conventional feature-based KD or self-distillation methods are inapplicable to the cross-architecture setting, while DUKD can still effectively improve the student models.
For instance, compared to the RCAN model trained from scratch, utilizing DUKD with an EDSR or SwinIR teacher model yields an increase in PSNR by 0.22dB at \texttimes 4 scale on Urban100 test set.

\noindent\textbf{Visual Comparison. }
In~\cref{fig:vis}, we compare the visual quality  of output images of the \texttimes 4 EDSR model trained by DUKD and other methods. To underscore the differences in detailed pattern and texture reconstruction, we took relatively small cropped portions and computed local PSNR metrics. Generally, a higher PSNR aligns with superior visual effect. For the reconstruction of textures (e.g. lines, edges, and complex patterns), the model trained with DUKD yields outputs that are both sharper and more similar to HR, indicating the superiority of DUKD.

\noindent\textbf{Experiment Results on Real-world SR task. } 
To test the performance of DUKD for real-world SR, we continue to train the PSNR-oriented student SwinIR models of \texttimes 4 scale by using the BSRGAN degradation model~\cite{zhang2021designing, liang2021swinir} on the DF2K dataset. The models are tested on three testing datasets: RealSR~\cite{cai2019toward}, DRealSR~\cite{wei2020component}, and OST300~\cite{wang2018recovering}. The non-reference image quality assessment (NIQE)~\cite{mittal2012making}
scores are
shown in Tab. \ref{tab:realworld-sr}. The model trained with DUKD produces lower NIQE scores and output images with more pleasing visual performance, as shown in supplementary material.

\subsection{Ablation Analysis}\label{sec: ablation}

\noindent\textbf{Impact of data upcycling and label consistency regularization.}~\cref{tab:ablation-du-lcr} shows the effect of the presented two modules, using EDSR baseline model (\#Channel=64, \#Block=16) distilled by our student model. Further,~\cref{tab:ablation-zi-zo} ablates the zoom-in~\faSearchPluss~and zoom-out~\faSearchMinuss~operations in data upcycling.
The result shows that adopting data upcycling and label consistency regularization could lead to significant performance improvement upon logits-KD, whether used individually or together. 
For example, simply upcycling data by zoom-in draws 0.31dB PSNR increment on Urban100 test set, and adding zoom-out andlabel consistency regularization yields an additional 0.16dB improvement. 

\begin{table}[]
\begin{minipage}[t]{0.55\linewidth}
    
\caption{Ablation study of data upcycling\\ and label consistency regularization. 
}\vspace{-0.5em} %
\label{tab:ablation-du-lcr}
\centering
\resizebox{0.85\columnwidth}{!}{%
\begin{tabular}{@{\hspace{1pt}}cc@{\hspace{6pt}}c@{\hspace{1pt}}}
\toprule
\multirow{2}{*}[-0.24em]{Data Upcycling} & \multirow{2}{0.2\columnwidth}[-0.24em]{\centering {\small Label consistency}} & Urban100 \\\cmidrule(lr){3-3}
                      &                         & PSNR/SSIM    \\ \midrule
\xmark & \xmark & 24.87 / 0.7431 \\
\cmark & \xmark & 25.20 / 0.7558 \\
\cmark & \cmark & 25.34 / 0.7609 \\ 
\bottomrule
\end{tabular}%
}
\end{minipage}
\quad
\begin{minipage}[t]{0.48\linewidth}
\raggedright
\caption{Ablation study of the\\ zoom in and zoom out operations. 
}\vspace{-0.5em}  %
\label{tab:ablation-zi-zo}
\resizebox{0.8\columnwidth}{!}{%
\begin{tabular}{@{\hspace{2pt}}ccc@{\hspace{2pt}}}
\toprule
Zoom In  & Zoom Out  & Urban100 \\ \cmidrule{3-3}
\faSearchPluss & \faSearchMinuss & PSNR / SSIM \\\midrule
\xmark                        & \xmark                         & 24.87 / 0.7431 \\
\cmark                        & \xmark                         & 25.18 / 0.7551 \\
\xmark                        & \cmark                         & 25.18 / 0.7552 \\
\cmark                        & \cmark                         & 25.20 / 0.7558 \\ 
\bottomrule
\end{tabular}%
}
\end{minipage}
\end{table}

\noindent\textbf{Integrate DUKD into other model compression methods.} We integrate DUKD with a SOTA SR network quantization method, Distribution-Aware Quantization (DAQ)~\cite{hong2022daq}, and use the full-precision model to supervise the quantized ones. \cref{fig:dukd_quantized} shows the PSNR of quantized \texttimes4 scale EDSR baseline models trained with and without KD, and the full results are provided in supplementary. It shows that DAQ with vanilla Logits-KD has barely effects on the model, while DUKD could improve the quantized model by a large margin. 
We also integrate DUKD with the FAKD method in~\cref{tab:fakd-dukd}. The resulting models outperform the ones trained by FAKD greatly. 
The results indicate that DUKD, as a data-centric approach, can be effectively aggregated with other model compression techniques.

\begin{table}[]
\begin{minipage}{0.5\linewidth}
    \centering
    \includegraphics[width=\columnwidth]{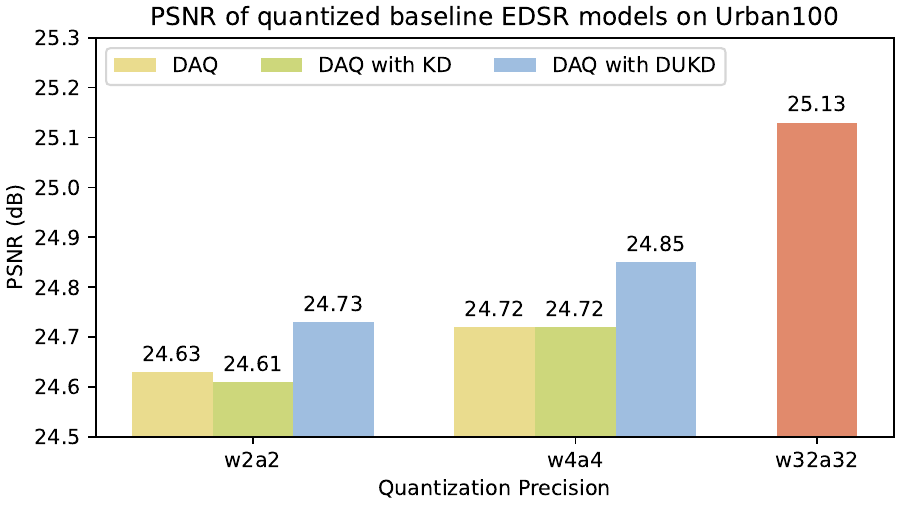}%
    \captionof{figure}{PSNR of quantized baseline EDSR model trained with and without KD. 
    }%
    \label{fig:dukd_quantized}
\end{minipage}
\quad\quad
\begin{minipage}[b]{0.4\linewidth}
\caption{Experiment results of combining DUKD and FAKD. 
}%
\centering
\label{tab:fakd-dukd}
\resizebox{\columnwidth}{!}{%
\begin{tabular}{@{\hspace{2pt}}llc@{\hspace{2pt}}}
\toprule
\multirow{2}{*}[-0.24em]{Model} & \multirow{2}{*}[-0.24em]{Method} & Urban100     \\ \cmidrule(lr){3-3} 
                       &                         & PSNR / SSIM    \\ \midrule
\multirow{4}{*}{EDSR}  & Logits KD               & 26.21 / 0.7897 \\
                       & FAKD                    & 26.18 / 0.7895 \\
                       & FAKD+DUKD               & 26.30 / 0.7930 \\
                       & DUKD                    & 26.45 / 0.7966 \\ 
\bottomrule
\end{tabular}%
}
\end{minipage}
\end{table}

\begin{table}[h]
\caption{Comparison with data expansion. DF2K denotes DIV2K+Flickr2K. 
}%
\label{tab:cmpr-dataexpansion}
\centering
\resizebox{0.8\columnwidth}{!}{%
\renewcommand{\arraystretch}{0.9}  %
\begin{tabular}{@{\hspace{1.5pt}}cccc@{\hspace{5pt}}c@{\hspace{5pt}}c@{\hspace{1.5pt}}}
\toprule
\multirow{2}{0.1\columnwidth}[-0.24em]{\centering Training set}  & \multirow{2}{*}[-0.24em]{\#Images} & \multirow{2}{0.15\columnwidth}[-0.24em]{\centering Training steps} & \multirow{2}{*}[-0.24em]{Method} & BSD100 & Urban100 \\ \cmidrule(lr){5-6} 
&         &    &    & PSNR/SSIM   & PSNR/SSIM   \\ \midrule
\multirow{2}{*}{DIV2K} & \multirow{2}{*}{800} & \multirow{2}{*}{$2.5\times10^5$} & Scratch & 27.57/0.7356 & 25.94/0.7809 \\
 &     &  & DUKD & {\bf 27.68/0.7390} & {\bf 26.32/0.7927} \\ \midrule
\multirow{2}{*}{DF2K} & \multirow{2}{*}{3450} & \multirow{2}{*}{$5\times10^5$} & Scratch & 27.62/0.7372 & 26.15/0.7872 \\
 &       &  & KD & 27.67/0.7390 & 26.31/0.7925 \\
\bottomrule
\end{tabular}%
}
\end{table}

\noindent\textbf{Comparison with data expansion} 
At the end of~\cref{sec: dukd}, we discussed the difference between DUKD and data augmentation/expansion. \cref{tab:cmpr-dataexpansion} compares DUKD with  training or KD with expanded data. 
We train the \texttimes 4 scale EDSR models on a much larger dataset (DF2K: DIV2K+Flickr2K~\cite{timofte2017ntire}, which contains 3450 images). 
The number of iterations is doubled for the larger training set since the previous configuration ($2.5\times10^5$) is insufficient for the models to converge. Except that the \texttimes 4 SR networks are not initialized with the \texttimes 2 ones in this comparison, the other settings of the training recipe are the same.
The result shows that DUKD is superior to training with more input data in terms of both efficiency and performance, indicating the necessity of data upcycling for KD.

\section{Conclusion}

In this work, we investigated the issues existed in KD for SR networks. Motivated by the findings, we present DUKD, a simple yet significant KD framework for SR, which outperforms existing methods and is applicable to a wide array of network architecture and SR tasks. 
Central to our approach is data upcycling, which facilitates the knowledge transfer from teacher to student model through the auxiliary upcycled data, without the interference of GT label. 
Besides, we realize label consistency regularization in KD for SR, which further bolsters the student model's generalization capabilities. 
Extensive experiments are conducted across various SR tasks, benchmark datasets and diverse network backbones, consistently showing the out-performance of DUKD and endorsing its robust and effective. 
In conclusion, the DUKD framework serves as a KD strategy with effective data utilization, harnessing the power of data upcycling and label consistency regularization to push the boundaries of SR model performance.
\newpage

\FloatBarrier

\bibliographystyle{splncs04}
\bibliography{main}

\end{document}